\documentclass[letterpaper, 10 pt, conference]{ieeeconf}  

\IEEEoverridecommandlockouts                              

\usepackage{cite}
\usepackage{amsmath,amssymb,amsfonts}
\usepackage{graphicx}
\usepackage{textcomp}
\usepackage{algorithm}
\usepackage{algpseudocode}
\usepackage{booktabs}
\usepackage{balance}
\usepackage{wrapfig}
\usepackage{xcolor}
\usepackage[table]{xcolor} 

\usepackage{makecell}
\usepackage{multirow}
\usepackage{longtable}

\usepackage[absolute,overlay]{textpos}
\title{\LARGE \bf
Receding-Horizon Next-Best-View Planner for Autonomous Leaf
Surface Reconstruction
}

\author{Arif Ahmed$^{1}$, Sajal K. Das$^{2}$, and Parikshit Maini$^{1}$
\thanks{$^1$ Systems and Algorithms for Robot Autonomy Lab (SARAL), Dept. of Computer Science and Engineering, University of Nevada, Reno, USA.}%
\thanks{$^2$ Department of Computer Science, Missouri University of Science and Technology, Missouri, USA.}
\thanks{Corresponding author: Parikshit Maini, email: pmaini@unr.edu. }%
\thanks{This material is based upon work supported by the US National Science Foundation under Grant Nos. 2544716, PFI-2431990, and AI-ENGAGE-2520346. Any opinions, findings, and conclusions or recommendations expressed in this material are those of the author(s) and do not necessarily reflect the views of the National Science Foundation.}
\thanks{\copyright  2026 IEEE. 
Personal use of this material is permitted. Permission from IEEE must be obtained for all other uses, including reprinting/republishing
this material for advertising or promotional purposes, collecting new collected works for resale or redistribution to servers or lists, or reuse of any copyrighted component of this work in other works}
}

\begin{document}

\begin{textblock}{100}(2,5)
\centering
{\small This paper has been accepted for publication in the IEEE/RSJ International Conference on Intelligent Robots and Systems (IROS 2026)
}
\end{textblock}

\maketitle
\thispagestyle{empty}
\pagestyle{empty}


\begin{abstract}

Accurate plant leaf modeling is fundamental to downstream
tasks such as plant growth monitoring, and phenotyping for
yield estimation. Autonomous robotic reconstruction for
large-scale field deployment must address limitations on
robot planning budget and computation resources while
optimizing viewpoint utility for leaf surface
reconstruction. Existing approaches either focus on rigid
objects, point-cloud coverage or plant reconstruction
without fully addressing the system limitations or
exploiting task-driven point cloud utility. In this work,
we study next-best-view (NBV) planning for leaf surface
reconstruction under travel constraints. We develop a novel
Centroid-based Information Gain (CIG) function that
measures the spatial distribution of observed points
relative to the centroid of the existing point cloud to
compute viewpoint utility. We also develop a
receding-horizon variant that reasons over future
viewpoints. To benchmark our work, we use the LAST-STRAW~\cite{uol2024laststraw} public dataset that includes point clouds of strawberry
plants over different growth stages and compare our method
with attention-driven NBV~\cite{burusa2024attention}  that uses a visibility-based
information gain approach. The proposed receding-horizon
approach consistently reduces surface reconstruction error
and improves geometric fidelity across multiple growth
stages, especially under increased inter-leaf occlusion.
Results demonstrate that our approach is able to visit
viewpoints that reduce surface reconstruction error and
improves reconstruction accuracy as compared to the
baseline by upto 10\%. 
\end{abstract}

\begin{keywords}
plant leaf reconstruction, view planning, next best view, leaf surface reconstruction, receding-horizon

\end{keywords}


\section{Introduction}
Accurate plant modeling is fundamental to downstream tasks such as phenotyping, growth monitoring, yield estimation, and precision agriculture \cite{roy2017active, zermas2017estimating, wei2024fast}.
Large-scale remote-sensing often lacks the resolution required for accurate plant and leaf-level 3D reconstruction \cite{zermas2017estimating}. Close-range specialized setups (such as a turntable) to scan individual plants can achieve higher fidelity \cite{wu2022miniaturized,liu2023fast,yuan2023field} but require manual plant placement and are impractical for large-scale field coverage. Ground-based mobile manipulators \cite{ yuan2023field,zhang2012application} and Unmanned Aerial Vehicles (UAVs) \cite{lin2021uav, zhang2024improved, tunca2024accurate} provide promising alternatives for field deployments, with each platform providing distinct trade-offs. Mobile manipulators may have restricted viewpoints due to limited access via aisles, but provide ease of control and robust data collection. Low-flying UAVs have better viewpoint access due to top-down visibility; however, flight controls are more involved. In this work, we consider a view-planning problem for UAV-based plant coverage with the goal to reconstruct all the leaves on the plant. 

\begin{figure}[t!]
    \centering
    \includegraphics[width=0.69\linewidth]{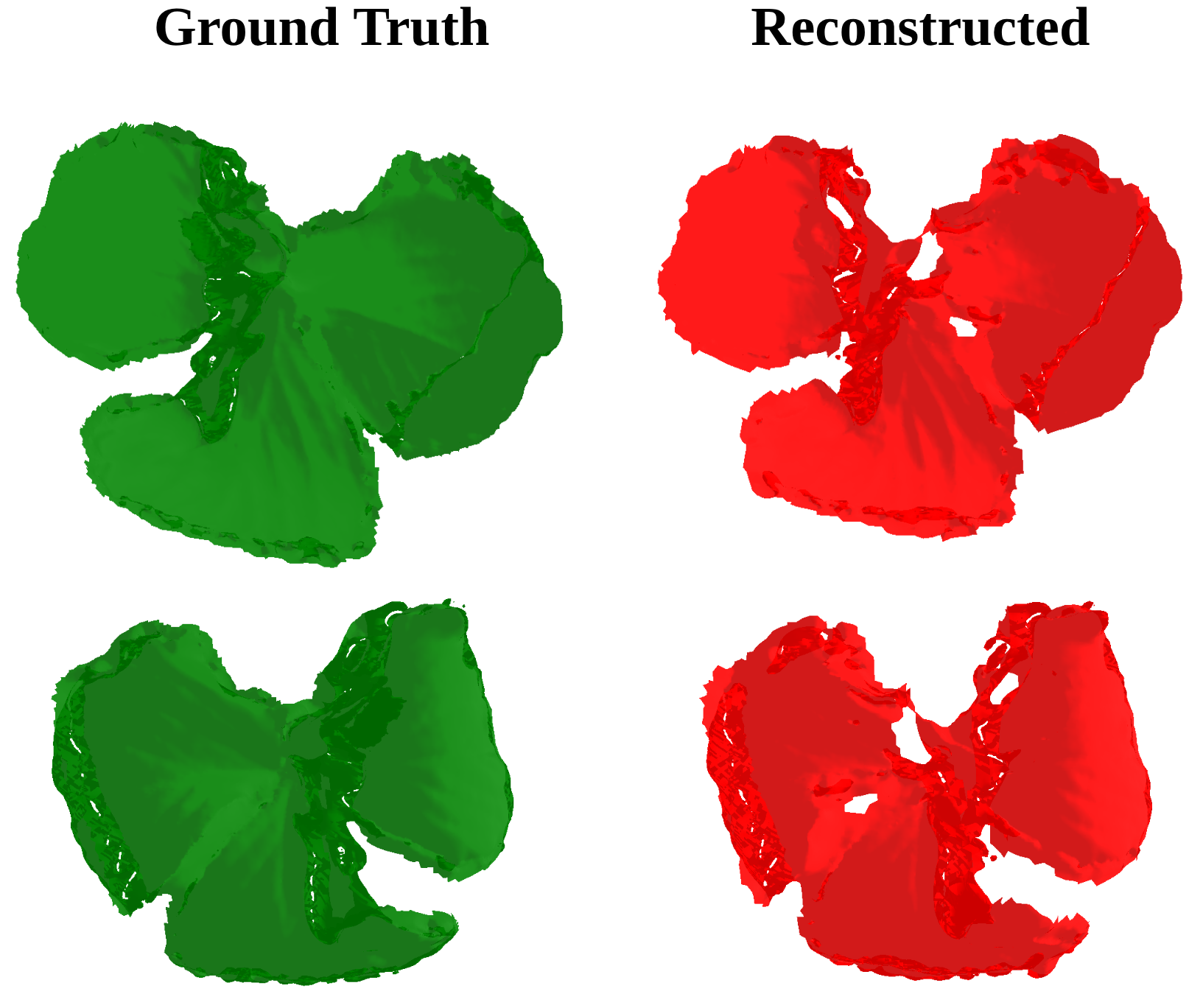}
    \caption{Leaf surface reconstruction of a trifoliate strawberry leaf using receding-horizon CIG-NBV (RH-CIG-NBV) planning for data collection.}
    \label{fig:existing}
\end{figure}

Next-best-view (NBV) planning methods have been proposed for 3D reconstruction tasks; however, their application to UAV-based plant scanning faces significant practical limitations~\cite{zeng2020pc, wu2022miniaturized, dhami2023pred, burusa2024attention, jia2025pb}. NBV methods typically estimate information gain per candidate viewpoint to select the most informative next view \cite{dhami2023pred}, but existing formulations were designed for rigid object reconstruction and do not account for specific geometric properties of leaf surfaces. Recent plant-based NBV planning approaches~\cite{burusa2024attention, burusa2024gradient} face two key limitations: (i) computational overhead and (ii) overlooking spatial geometry.  First, they rely on a computationally intensive pipeline that converts RGB-D data to point clouds, voxelizes and constructs OctoMaps, performs ray-tracing, and computes Shannon entropy for information gain estimation. Such pipelines are very expensive for resource-constrained platforms like UAVs that operate under strict travel and computational budgets. Second, current information gain functions treat all visible points equally, assigning the same weights to viewpoints that observe spatially clustered points as to those revealing distributed coverage across leaf surfaces. Existing information gain functions treat all visible points equally, although their impact on surface reconstruction varies greatly~\cite{burusa2024attention, burusa2024gradient}. 
In this work, we make the following contributions:
\begin{itemize}
    \item A \textbf{C}entroid-based \textbf{I}nformation \textbf{G}ain (CIG) function that measures the spatial distribution of observed points relative to the centroid of the existing point cloud, explicitly rewarding viewpoints that observe points farther from already viewed regions and maximizing spatially distributed observations across leaf surfaces (sample reconstruction in Fig. \ref{fig:existing}).
    \item A receding-horizon view planning formulation under explicit travel budget constraints that plans a sequence of viewpoints within a finite horizon window rather than greedily selecting a single next-best-view at each step.
    \item Comprehensive evaluation on the LAST-STRAW dataset~\cite{uol2024laststraw} across 14 growth stages from seedling to fruiting maturity, demonstrating consistent improvements in per-leaf point cloud coverage and leaf surface reconstruction quality, and showing that downstream surface reconstruction metrics reveal planning strategy differences that point cloud coverage metrics alone cannot capture.
\end{itemize}

The remainder of the paper is organized as follows. Section II reviews related work on NBV planning and plant reconstruction. Section III explains the visibility model, followed by problem formulation in Section IV. Section  V describes our CIG-based view planning approach, Section VI outlines the experimental setup, and Section VII presents the evaluation metrics and results across all growth stages and planning strategies. Finally, Section VIII concludes with directions for future work.

\section{Background and Related Work}

View planning aims to find a set of viewpoints to achieve a specific goal such as object or environment reconstruction. In most cases, such tasks involve constraints such as robot mobility, sensor limitations, and environmental complexity. However, it might not always be possible to thoroughly scan the environment using onboard sensors for real-time constraints. In such cases, robots incrementally move the sensors to determine a single next best view (NBV) instead of a sequence of viewpoints, to get the most new information from the environment. 

\subsection{View Planning for Object Reconstruction}
Classical NBV approaches evaluate candidate viewpoints by estimating new observable regions and selecting the view that maximizes information gain \cite{ dhami2024map}. 
Dhami et al. \cite{dhami2023pred} sample viewpoints along concentric circles around the object and estimate visibility using the Hidden Point Removal (HPR) operator \cite{katz2007direct}.  
The candidate revealing the largest number of new points is selected, subject to the closest candidate viewpoint (lowest motion cost) that provides sufficient information gain via RRT-Connect~\cite{kuffner2000rrt}.
Their framework integrates  PoinTr \cite{yu2021pointr}, transformer-based shape completion  model fine-tuned using curriculum framework\cite{bengio2009curriculum} on ShapeNet 
to predict unseen geometry from partial observations. While completion improves visibility prediction, the approach introduces additional model dependency and computational overhead.
Recently, Jia et al.~\cite{jia2025pb} proposed a voxel-based NBV planning strategy inspired by 3D Gaussian Splatting.
To reduce computational cost, they replace traditional ray-casting with projection-based visibility evaluation. They evaluate candidate viewpoints based on projected observation quality from classified frontier and occupied voxels.
These methods are effective for rigid objects where information gain implicitly treats all visible elements equally, and such formulations do not explicitly account for geometric structures with redundant observations for leaves.

\subsection{View Planning for Plant Reconstruction}
In plant phenotyping, many reconstruction approaches rely on controlled environments and turntable setups~\cite{wei2024fast}, where a static plant is rotated under uniform, stable LED lighting with diffused illumination and black-background enclosures to minimize natural light inference and improve depth sensing accuracy. 
In \cite{wu2022miniaturized}, the researchers proposed a platform for phenotyping individual plants to estimate traits such as plant height and leaf area, using multiview stereo 3D reconstruction.
The plant remains stationary while the camera arms rotate $360^\circ$ around it to capture images from multiple viewpoints. To minimize light and wind effects, they include a wind-shield and LED lighting. 
Importantly, authors ensured the high-quality image by setting up controlled imaging conditions to avoid light and wind effects.  
Similarly, \cite{liu2023fast} reconstructed a potted peanut plant using two symmetrically placed RGB-D sensors, aligning and merging dual point clouds followed by ROI-based PassThrough filtering and statistical outlier removal. While effective, such systems depend on constrained acquisition setups and predefined filtering parameters. Zermas et al. \cite{zermas2017estimating} reconstruct corn plants using circular handheld camera scanning, applying VisualSFM for 3D reconstruction,   though this approach lacks strategic viewpoint selection.

In contrast, several works address agricultural robotics problems for fruit counting, fruit shape, plant parts, or plant reconstruction using a mobile manipulator~\cite{menon2023nbv, burusa2024gradient, burusa2024attention, isaacjose2025iros}.
For example, 
Roy et al.  \cite{roy2017active} proposed a view planning method to count apples in orchards using geometric information.
Recently, Li et al. \cite{li2025semp} proposed SemP-NBV, which addressed Pred-NBV's \cite{dhami2023pred} fixed sampling through PCA-based adaptive candidate generation and gradient-optimized HPR radius. Despite these improvements, the framework relies on a transformer-based shape completion model AdaPointTr, 
which introduces substantial computational overhead for resource-constrained UAV deployment.

More recently, Burusa et al. \cite{burusa2024attention}  introduced attention-driven NBV planning that focuses on task-relevant regions (e.g., leaves) rather than treating all plant parts equally. Their method integrates RGB-D measurements into probabilistic OctoMaps and evaluates candidate viewpoints through ray-tracing-based information gain constrained to predefined bounding boxes around the target ROI. 
Despite improving task relevance, the method lacks spatial awareness in information gain computation. The volumetric formulation treats voxel visibility as binary (visible or not), ignoring geometric distribution. The distinction between redundant clustered observations and valuable distributed coverage is lost.
Later, they proposed Gradient-NBV \cite{burusa2024gradient}, a local planner that avoids global sampling of candidate viewpoints and instead uses gradient ascent with differentiable ray sampling to move the camera in the direction that maximizes semantic information gain for a single ROI. 

Despite these advances, existing plant-oriented NBV planning approaches face two key limitations. First, volumetric mapping and per-view ray tracing introduce substantial computational overhead, limiting real-time deployment on resource-constrained platforms such as UAVs.
Second, entropy-based information metrics treat all visible elements independently and equally without explicitly rewarding  complementary observations, 
i.e., viewpoints that reveal previously unseen regions of the same leaf surface and contribute additional geometric structure for more complete reconstruction.
\section{ Visibility Model}
\label{sec:hpr}
\begin{figure}[!htbp]
    \centering
    \includegraphics[width=0.65\linewidth]{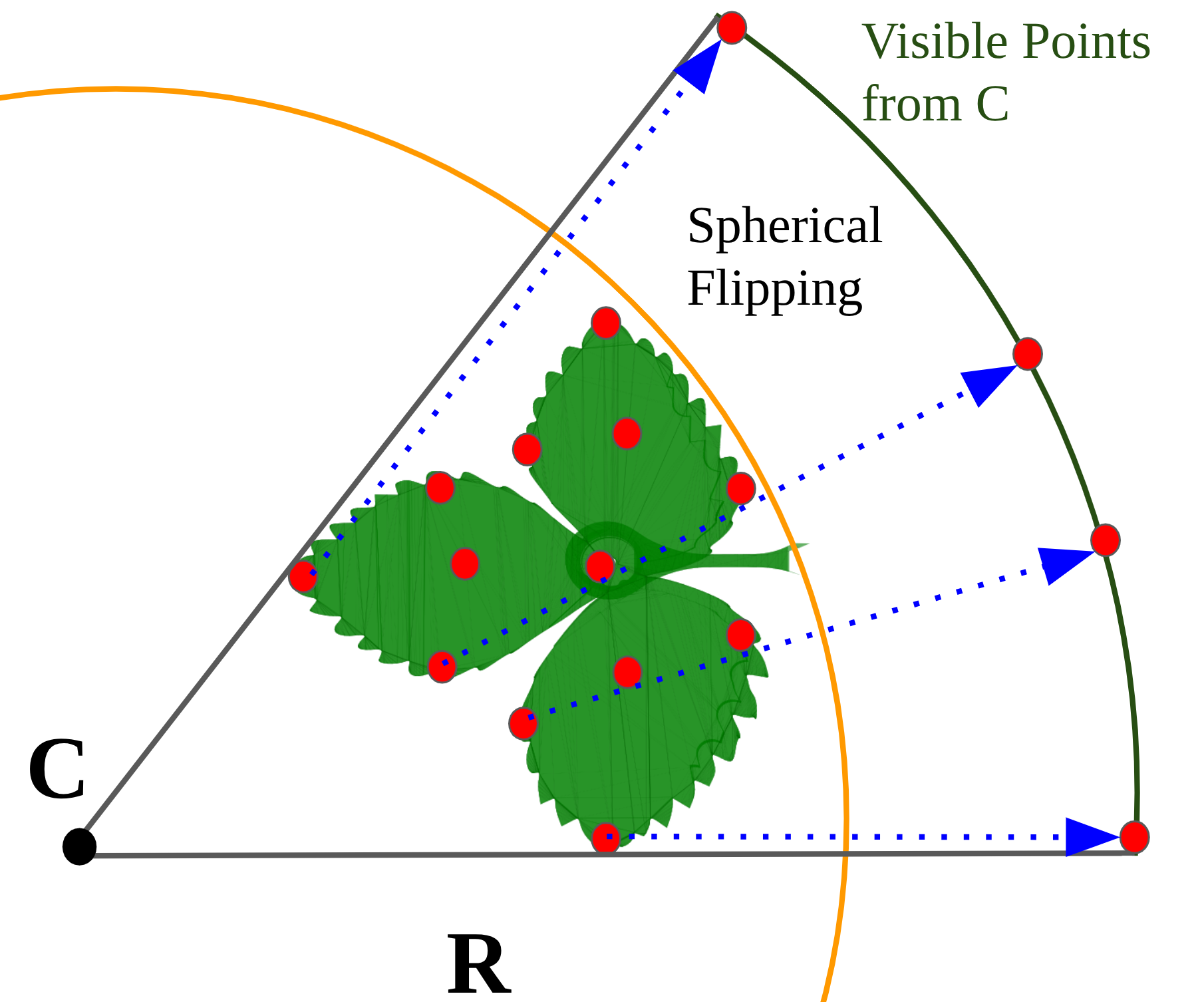}
    \caption{Determine visible points using HPR operator (illustrated in 2D). }
    \label{fig:hpr_explained}
\end{figure}

Determining points on the plant that are visible from a given UAV pose is central to informative view planning.
Ray-tracing methods need a voxelized point cloud or need to reconstruct the surfaces \cite{burusa2024attention, katz2007direct} to compute visibility.
Therefore, we adopt Hidden Point Removal (HPR) \cite{katz2007direct} as our visibility model, which computes the visible subset of a point cloud from a specific camera viewpoint without mesh reconstruction or normal estimation. 
HPR operates in two steps. First, each point $p_i$ is reflected through a sphere of radius $R$ centered at the camera location $C$ via spherical flipping:
\( \hat{p_i} = f(p_i) = p_i + 2 (R- \|p_i\|) \frac{p_i}{\|p_i\|}\).
Second, the convex hull of the transformed point cloud $\hat{P} \cup C$ is computed. A point $p_i$ is considered visible from $C$ if and only if its flipped counterpart $\hat{p_i}$ lies on the convex hull, as illustrated in Fig~\ref{fig:hpr_explained}.

  Given the visible point indices returned by HPR at viewpoint $v$, we extract the visible subset for each leaf instance $l$:
\( \mathcal{V}(v,l) = \{ i: V_{v,i} = 1, l_i = l \}\), where $V_{v,i} =1$ indicates $p_i$ is visible from viewpoint $v$.
\section{Problem Formulation}
To set up our simulation, we consider a UAV inspecting a plant to maximize observed leaf surface coverage across all leaf instances. 
 The UAV carries a depth sensor and visits a sequence of viewpoints surrounding a plant, collecting surface observations at each position. 
 For planning purposes, we  simulate this process over a 3D point cloud reconstruction of the plant, where each point carries a leaf instance label 
 (see Sec~\ref{sec:pointnet}) and visibility from each candidate viewpoint is estimated via HPR (as discussed in Sec.~\ref{sec:hpr}).

\textbf{Problem Definition:}
Given a 3D plant point cloud $\mathcal{P}= \{(p_i, l_i)\}_{i=1}^N$, where $p_i \in \mathbb{R}^3$ and $l_i$ is the leaf instance ID, with $L$ leaf instances, a candidate viewpoint set $ \mathcal{C} = \{v_1,v_2, ..,v_M\}$, and a flight budget $\mathbf{B}$,  the planner selects a closed-loop view 
sequence $\mathcal{S} = [s_1, s_2, ...,s_k] \in ~\mathcal{C}$
that maximizes 
observed leaf surface coverage across all $L$ leaf instances.
We define the cost of a closed-loop view sequence $S = [s_1, s_2, ..., s_k]$ returning to the origin as:
\( Cost(\mathcal{S}) = \sum_{t=1}^{k-1} \| v_{s_t} - v_{s_{t+1}}\|_2 ~+ ~\|v_{s_k} - v_{s_1}\|_2 \).

\noindent The planner solves:
\( \max_\mathcal{S} \sum_{l=1}^L\left| \bigcup_{t=1}^k \{i: l_i=l, \mathcal{V}_{s_t,i}=1\} \right|\) subject to \( Cost(\mathcal{S})\le\mathbf{B}\ ,  k\le M\),  
where $l_i$ identifies the leaf instance of point $i$, $\mathcal{V}_{s_t,i}=1$ indicates visibility from $v_{s_t}$ via HPR, and $\mathcal{S}$ contains no repeated viewpoints.
We address this in Sec~\ref{sec:cig} by proposing a centroid-based information gain function, evaluated under both a greedy NBV and a receding horizon planning strategy. 

 \section{Method}
We present a view planning framework for UAV-based plant leaf surface reconstruction that combines a novel Centroid-based Information Gain (CIG)  with a receding-horizon planning strategy ~\cite{bircher2016receding, bircher2018receding}. 
We directly compute visibility from the plant point cloud using HPR (see Sec.~\ref{sec:hpr}), eliminating voxelization and ray tracing. 
CIG measures the spatial distribution of observed points relative to the centroid of already-viewed leaf surface, rewarding viewpoints that reveal spatially distant regions over redundant ones. 
We evaluate CIG under both a greedy NBV and a receding horizon formulation (horizon window of 3) under the travel budget $B$.

 \subsection{Candidate Viewpoints Set}
 \label{sec:candidate_viewpoints_set}
\begin{figure}[!htbp]
    \centering
    \includegraphics[trim=0pt 10pt 0pt 10pt, clip, width=0.99\linewidth]{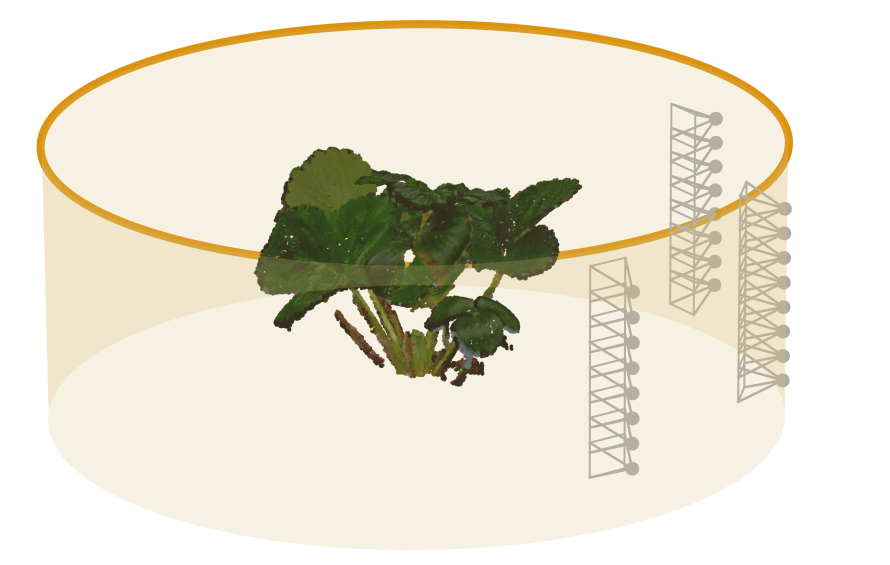}
    \caption{Candidate UAV viewpoints (camera poses) sampled on a quarter-cylindrical surface around the plant, facing the plant center. 
    }
    \label{fig:candidate_pose}
\end{figure}

 To generate candidate viewpoints, we consider a cylindrical surface around the plant, with the radius determined by the plant type and growth stage. We generate $M$ candidate viewpoints uniformly distributed on the cylinder facing the plant center  
 (Fig.~\ref{fig:candidate_pose}). 
 We select the viewpoints to provide coverage along the height of the plant within one quarter of the cylinder, representing the access restrictions due to the field layout.

\subsection{ Point Cloud Semantic Labels}
\label{sec:pointnet}
We incorporate semantic information at the instance level for targeted leaf perception, inspired by the targeted perception framework in
\cite{burusa2024gradient}.
Each point $p_i$ in the scanned plant point cloud receives a leaf instance ID $l_i \in \{ 0, 1, ...,(L-1)\} $.
We use these labels throughout planning to extract per-leaf visible point sets $\mathcal{V}(v,l)$ using the HPR operator.

\subsection{Proposed Centroid-based Information Gain (CIG)}
\label{sec:cig}
\begin{figure}[!htbp]
    \centering
    \includegraphics[width=0.99\linewidth , trim=0pt 20pt 0pt 25pt, clip]{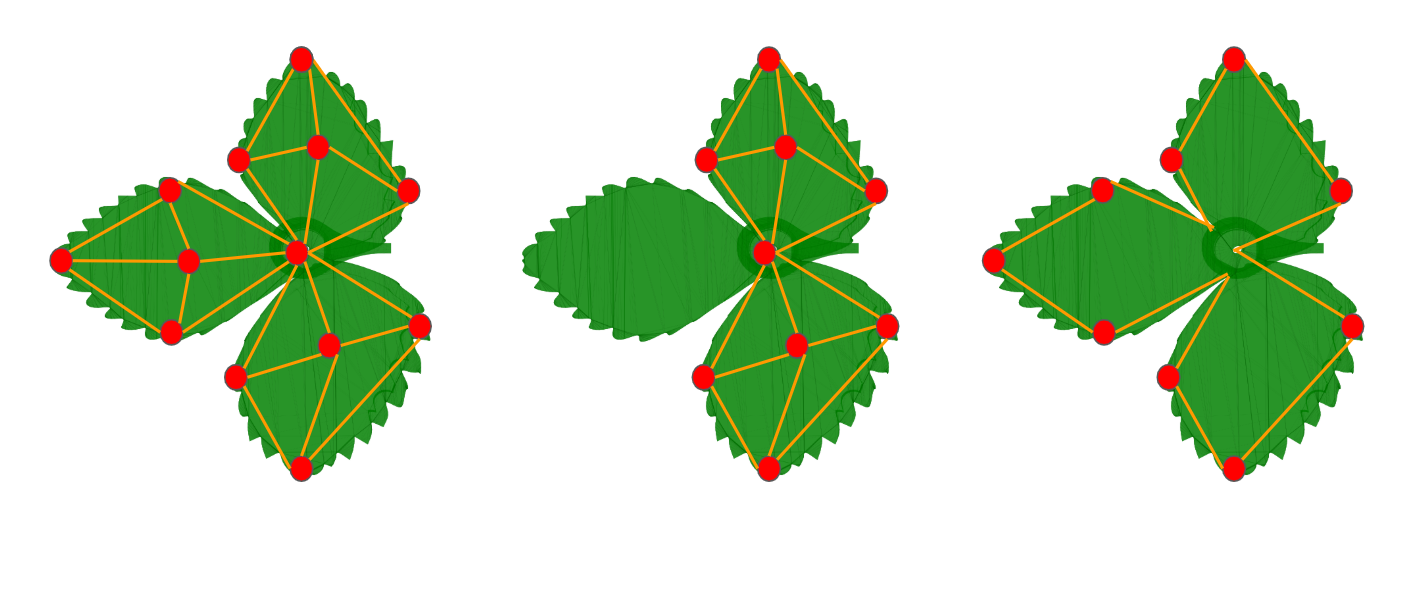}
    \caption{
    Limitations of existing IG functions are illustrated in 2D. Ground truth leaf (Left), nine points with partial coverage (two-thirds) (Center), and nine points with complete coverage (Right). Existing methods treat both equally; our proposed CIG differentiates them based on spatial distribution. 
    }
    \label{fig:cig_concept}
\end{figure}

Existing IG functions treat all visible points equally, assigning the same gain regardless of whether newly observed points are spatially clustered or distributed across the leaf surface, as illustrated in Fig.~\ref{fig:cig_concept}, where existing methods fail to differentiate the partial and complete coverage.
To address this limitation, we introduce a Centroid-based Information Gain (CIG) function that measures the spatial distribution of observed points relative to the centroid of existing points. 
Unlike existing IG functions, CIG explicitly rewards viewpoints that observe points farther from the centroid of already-viewed plant parts, thereby maximizing spatially distributed observations across leaf surfaces (Fig.~\ref{fig:cig_concept}).
This formulation directly addresses the spatial coverage limitation,  where viewpoints revealing spatially distant leaf regions assign higher information gain than those observing clustered, redundant regions.

At each planning step $t$, we maintain the set of already observed points for each leaf instance $l$:
\( \mathcal{O}_t^l = \bigcup_{\tau=1}^t \mathcal{V}_{s_\tau , l} \) 
 with centroid: 
\(\text{} \mu_t^l \frac{1}{|\mathcal{O}_t^l|} \sum_{i\in\mathcal{O}_t^l} p_i\). 
For a candidate viewpoint $v$, we compute the distance of each newly visible leaf point from the per-leaf centroid:
\( d_i^l = \| p_i - \mu_t^l\|_2 , \forall_{p_i} \in \mathcal{V}_{v,l} \).
The CIG for candidate viewpoint $v$ is then:
\begin{equation}
    \label{eq:cig}
CIG(v) = \sum_{l=1}^L \sum_{i\in \mathcal{V}_{v,l}} exp \left(\frac{d_i^l}{D^{*l}_{robust}} \right) 
\end{equation}
\( \text{where, } D^*_{robust} = \text{percentile}_{95}(\{d_i\}_{i=1}^{|\mathcal{O}_t^l|} \). 
This formulation assigns a higher gain to viewpoints that observe points far from the centroid of previously covered points.

\subsection{Receding-Horizon NBV Planning}
We formulate view planning as a receding-horizon optimization problem under explicit budget constraints, where we plan a sequence of viewpoints within a finite horizon window $(w=3)$ rather than greedily selecting a single NBV. 
For each candidate viewpoint, we apply HPR to determine visible points and compute CIG. 
We evaluate all candidate sequences within the horizon window, select the path that maximizes CIG while satisfying the remaining budget, execute the first viewpoint in the planned sequence, and repeat the process until the budget is exhausted.
Formally, at each step $t$ the planner solves:
\begin{equation}
    \label{eq:rh-nbv}
    \mathcal{S}_t^* = \arg\max_{\mathcal{S}_t} \sum_{v \in \mathcal{S}_t} CIG(v) \text{ s.t. } Cost(\mathcal{S}_t) \le B_t
\end{equation}

where $B_t = B - Cost(\mathcal{S}_{1:t})$ is the remaining budget and $|\mathcal{S}_t| \le w$. 
The first viewpoint $s_t^*$ is executed, observed points are added to $\mathcal{O}^l_t$ for all $l$, and the process repeats. 
We describe the full procedure as summarized in Algorithm~\ref{alg:rhnbv}.

\begin{algorithm}[htbp]
\caption{RH-NBV-CIG}
\label{alg:rhnbv}
\begin{algorithmic}[1]
\State \textbf{Input:} $\mathcal{P}$, $\mathcal{C}$, $B$, $w$ 
\State \textbf{Output:} View sequence $\mathcal{S}$ 
\State $\mathcal{S} \leftarrow [s_1]$,  $B_t \leftarrow B$, $\mathcal{O}_0^l \leftarrow \emptyset$ $\forall l$
\While{$B_t>0$}
    \State Generate all sequences $\{\mathcal{S}_t\}$  with length $\leq w$ 
    \For{each candidate sequence $\mathcal{S}_t$}
        \If{$Cost(\mathcal{S}_t) \leq B_t$}
            \For{each $v \in \mathcal{S}_t$}
                \State apply HPR to $\mathcal{P}$ from $v$ $\rightarrow$ extract $\mathcal{V}_{v, l}$ $\forall$ $l$
                \State Compute $CIG(v)$ 
            \EndFor 
        \EndIf 
    \EndFor 
    \State $\mathcal{S}^*_t  \leftarrow \arg\max_{\mathcal{S}_t} \sum_{ v \in \mathcal{S}_t} CIG(v) $
    \State Execute $s_t^* \leftarrow$ first viewpoint of $\mathcal{S}^*_t$
    \State $\mathcal{O}_t^l \leftarrow \mathcal{O}_{t-1}^l \cup \mathcal{V}_{s_t^*, l}$ $\forall l$
    \State $ B_t \leftarrow B_t - \|v_{s_{t-1}} - v_{s_t^*} \|_2$
    \State $\mathcal{S} \leftarrow \mathcal{S} \cup[s_t^*]$
\EndWhile 

\State \Return $\mathcal{S}$
\end{algorithmic}
\end{algorithm}

\section{Experimental Setup}
This section describes the dataset, implementation details for candidate viewpoint generation, budget generation, visibility operator, and benchmark method to evaluate our work. 

\textbf{Dataset. }
We use the LAST-STRAW dataset \cite{uol2024laststraw}, which provides high-resolution 3D point clouds of two strawberry plants, scanned across 14 time steps spanning 11 weeks, covering different developmental stages. 
We evaluate on plants A2 and B1, which provide temporally consistent per-leaf instance annotations across all time steps i.e., growth stages.

\textbf{Travel Budget. }
We generate a circular trajectory around the target plant (see Fig.~\ref{fig:candidate_pose}). 
The circumference of this trajectory defines the travel budget,  $B_1$. We further define $B_2= B_1 \times 2 $, and $B_3 = B_1 \times 3$. 
We impose these budget levels as travel constraints in all planning experiments.

\textbf{Implementation Details. }
For computational efficiency and fair comparison across all planning methods, we generate 24 candidate viewpoints in a quarter-cylindrical sector ($\mathrm{azimuth} \in [0^\circ, 90^\circ]$, 3 azimuthal columns $\times$ 8 elevation rows).
The cylinder radius is set to $0.6$ times the maximum of the plant's XY bounding box dimensions, with the plant's Z-range as the height.
We apply HPR \cite{katz2007direct} (radius $r = 50.0$)  using Open3D  to determine point visibility from each candidate viewpoint.
For leaf surface reconstruction, we use Poisson reconstruction \cite{kazhdan2006poisson} (depth=9) with $5\%$ density-based filtering for noise removal implemented in Open3D.

\textbf{Baseline. }
We compare our method with Attention-driven NBV (Attn-NBV) as a baseline that implements the attention mechanism from \cite{burusa2024attention}, applying coverage-count information gain (IG) exclusively to leaf points only.
We evaluate two variants of our approach. Greedy CIG NBV (CIG-NBV) uses our centroid-based information gain (Eq.\ref{eq:cig}) with greedy selection, demonstrating the benefit of geometric IG over count-based methods. 
RH-CIG NBV applies receding horizon planning with horizon window $w=3$  using CIG (Eq.\ref{eq:cig}), representing our complete proposed method.%
\section{Results and Discussion}
In this section, we first describe evaluation metrics for point cloud coverage and leaf surface reconstruction.
Next, we present our experimental results and qualitative analysis. We evaluate our approach on our selected dataset using an identical viewpoint set and budget constraints for fair comparison. We use Attention-NBV \cite{burusa2024attention} as our primary benchmark. 

We evaluate results at three budget levels ($B_1, B_2, B_3 $) to analyze the coverage-budget relationship for resource-constrained UAV deployment. 
We report both point cloud coverage (Table~\ref{tab:f1-score}) and leaf surface reconstruction quality (Table~\ref{tab:qualitative}) to validate that coverage improvements correspond to geometric fidelity.

\subsection{Evaluation Metrics} 

\subsubsection{Leaf Point Cloud Coverage}
We evaluate view planning performance using a cloud-to-cloud comparison between the ground truth $G$ and the reconstructed point cloud $R$
using F1-Score \cite{knapitsch2017tanks}. 
We define $G$ as the union of the point clouds visible from each candidate viewpoint.  
For each reconstructed point $r \in R$, we compute its nearest-neighbor distance to the reference:
\( d_r = \min_{g \in G} \| r-g\| \).
Given a distance tolerance $d$, the reconstruction \textit{precision} %
is:
\( P(d) = \frac{1}{|R|} \sum_{r \in R} \mathbf{1}(d_r<d) \). 
For each reference point $g \in G$, its distance to the reconstruction is:
\( d_g = \min_{r \in R} \| g-r\| \). 
The reconstruction \textit{recall} 
is:
\( R(d) = \frac{1}{|G|} \sum_{g \in G} \mathbf{1}(d_g<d) \). 
To jointly assess precision and recall, we use the \textit{F1-score}, computed as the harmonic mean of precision and recall.

\begin{table*}[]
    \centering
    \caption{F1-Score Performance of Our Planning Methods CIG and RH-CIG  with respect to Attention Benchmark}
\renewcommand{\arraystretch}{1.2} 
\begin{tabular}{l|ccc|ccc|ccc}
\hline
\hline 
& \multicolumn{3}{c|}{ Travel Budget: $B_1$  } 
& \multicolumn{3}{c|}{Travel Budget: $B_2$ }
& \multicolumn{3}{c}{Travel Budget: $B_3$  } \\

\cline{2-10}
\multicolumn{1}{c|}{\multirow{2}{*}[10pt]{
\textbf{Plant ID }}
} & \textbf{Benchmark~\cite{burusa2024attention}} & \textbf{CIG} &  \textbf{RH-CIG} & \textbf{Benchmark~\cite{burusa2024attention}}  & \textbf{CIG}  & \textbf{RH-CIG}  & \textbf{Benchmark~\cite{burusa2024attention}}  & \textbf{CIG} & \textbf{RH-CIG}   \\
\hline 

A2\_20220512\_a (A21) & \underline{97.93} & \underline{97.93} & 97.16 & 99.52 & 99.58 & \underline{99.79} & 99.93 & 99.99 & \textbf{\underline{100.0}} \\
A2\_20220519\_a (A22)& 95.13 & 95.13 & \underline{98.77} & \underline{99.38} & \underline{99.38 }& 98.28 & 99.88 & 99.81 & \textbf{\underline{99.91}} \\
A2\_20220525\_a (A23) & 94.11 & 94.11 & \underline{97.50} & 99.21 & 99.17 & \underline{99.39} & 99.81 & 99.78 & \textbf{\underline{99.84}} \\
A2\_20220531\_a (A24) & 92.42 & 92.42 & \underline{96.19} & 98.46 & 98.46 & \underline{99.09} & 99.50 & 99.61 & \textbf{\underline{99.69}} \\
A2\_20220608\_a (A25) & 94.72 & 92.70 & \underline{97.43} & 98.55 & 99.07 & \underline{99.11} & 99.74 & \textbf{\underline{99.80}} & 99.76 \\
A2\_20220616\_a (A26)& 95.46 & 95.48 & \underline{96.15} & 99.15 & 98.90 & \underline{99.35} & \textbf{\underline{99.74}} & 99.67 & 99.70 \\
A2\_20220620\_a (A27)& 95.37 & 95.55 & \underline{95.70} & 99.11 & \underline{99.24} & 99.15 & 99.76 & \textbf{\underline{99.85}} & 99.84 \\
A2\_20220623\_a (A28) & \underline{95.59} & \underline{95.59} & 95.45 & 99.01 & \underline{99.10} & 99.08 & 99.71 & 99.76 & \textbf{\underline{99.82}} \\
A2\_20220627\_a (A29) & 95.45 & 95.45 & \underline{95.88} & 99.07 & \underline{99.17} & 99.13 & 99.82 & 99.82 & \textbf{\underline{99.83}} \\
A2\_20220629\_a (A2A) & 95.55 & 95.55 & \underline{96.89} & \underline{99.20} & \underline{99.20} & 98.96 & 99.70 & 99.77 & \textbf{\underline{99.82}} \\
A2\_20220707\_a (A2B) & 94.63 & 94.04 & \underline{95.19} & 98.94 & 99.08 & \underline{99.17} & \textbf{\underline{99.96}} & 99.95 & 99.74 \\
A2\_20220715\_a (A2C) & 94.82 & 94.82 & \underline{97.88} & 98.83 & 98.77 & \underline{99.32} & 99.74 & 99.62 & \textbf{\underline{99.90}} \\
A2\_20220721\_a (A2D) & 94.14 & 94.14 & \underline{96.69} & 98.66 & 98.68 & \underline{99.06} & 99.52 & \textbf{\underline{99.79}} & 99.75 \\
A2\_20220729\_a (A2E) & 95.10 & \underline{96.78} & 96.56 & 98.80 & 99.37 & \underline{99.52} & 99.88 & \textbf{\underline{99.98}} & 99.42 \\
\hline 
\multicolumn{1}{r|}{\textbf{Mean}}  & 95.03 &	94.98 &	\underline{96.67} & 98.99 &	99.08 & \underline{99.17} &	99.76 &	\underline{\textbf{99.80}} &	99.79 \\

\hline 
\hline 
B1\_20220513\_a (B11) & \underline{95.47} & 94.92 & 95.13 & \underline{99.39} & 98.80 & 99.34 & 99.87 & 99.78 & \textbf{\underline{99.92}} \\
B1\_20220520\_a (B12) & 95.03 & 95.03 & \underline{95.26} & 98.80 & 98.87 & \underline{99.33} & 99.79& 99.79 & \textbf{\underline{99.84}} \\
B1\_20220526\_a (B13) & 93.45 & 93.45 & \underline{97.17} & 99.01 & 99.14 & \underline{99.49} & \textbf{\underline{99.84}} & \textbf{\underline{99.84}} & 99.78 \\
B1\_20220531\_a (B14) & 95.97 & 95.97 & \underline{96.61} & 98.84 & 98.84 & \underline{98.94} & \textbf{\underline{99.77}}& \textbf{\underline{99.77}} & 99.17 \\
B1\_20220609\_a (B15) & 92.30 & 92.30 & \underline{95.78} & 98.59 & 98.72 & \underline{99.48} & 99.78 & 99.92 & \textbf{\underline{100.0}} \\
B1\_20220617\_a (B16)  & 93.94 & 93.94 & \underline{94.63} & 98.76 & 98.76 & \underline{99.04} & \textbf{\underline{99.94}} & 99.89 & 99.89 \\
B1\_20220620\_a (B17) & 90.53 & 92.55 & \underline{93.23} & 97.64 & 98.10 & \underline{98.68} & 99.65 & 99.65 & 99.05 \\
B1\_20220623\_a (B18) & 89.14 & 89.14 & \underline{89.56} & 97.35 & 97.35 & \underline{98.27} & 99.04 & 99.04 & \textbf{\underline{99.39}} \\
B1\_20220627\_a (B19) & \underline{ 91.75} & \underline{91.75} & 87.57 & 98.52 & 98.50 & \underline{98.74} & 99.68 & 99.52 & \textbf{\underline{99.87}} \\ 
B1\_20220630\_a (B1A) & 89.60 & 89.60 & \underline{93.19} & 97.63 & 98.07 & \underline{98.80} & 99.42 & 99.57 & \textbf{\underline{99.61}} \\
B1\_20220707\_a (B1B) & 89.42 & 91.27 & \underline{92.49} & 97.32 & 97.36 & \underline{98.09} & 99.01 & 99.16 & \textbf{\underline{99.47}} \\
B1\_20220715\_a (B1C) & 93.16 & 93.16 & \underline{96.27} & \underline{98.54} & 98.39 & 98.46 & \textbf{\underline{99.62}} & 99.40 & 99.57 \\
B1\_20220721\_a (B1D) & \underline{89.95} & \underline{89.95} & 89.91 & 97.75 & 97.47 & \underline{97.83} & 99.54 & 99.74 & \textbf{\underline{99.80}} \\
B1\_20220729\_a (B1E) & \underline{92.75} & \underline{92.75} & 90.82 & 97.59 & 97.28 & \underline{98.26} & 99.43 & 99.69 & \textbf{\underline{99.76}} \\
\hline
\multicolumn{1}{r|}{\textbf{Mean}} & 92.32 &	92.56 &	\underline{93.40} &	98.27 & 	98.26 &	\underline{98.77} &	99.60 &	99.63 &	\underline{\textbf{99.65}} \\

\hline 
\hline 
\addlinespace
\multicolumn{10}{l}{ \underline{Underline}: best within travel budget category; \textbf{bold}: best across travel budget levels $B_1$, $B_2$, and $B_3$. 
}
 
\end{tabular}
\label{tab:f1-score}

\end{table*}
\subsubsection{Leaf Surface Reconstruction Quality}
\label{sec:leaf_surface_metrics}
To reconstruct the leaf surface for both reconstructed points after view planning and ground-truth, we apply the Poisson surface reconstruction~\cite{kazhdan2006poisson}.
To capture surface accuracy, orientation fidelity, distance between surface points, and surface area accuracy, we use Hausdorff distance (HD)\cite{rogers1998hausdorff,aspert2002mesh}, normal consistency (NC)\cite{mescheder2019occupancy}, Root-Mean-Square (RMS), and relative error (RE).

\textbf{ Hausdorff Distance (HD): }The distance between a point $p$ on the reconstructed surface $M_i$ to the ground truth surface $M_t$, 
\(
d(p, M_t) = \min_{p_t \in M_t} \| p - p_t \|_2
\),
where $\| \cdot \|_2$ denotes the Euclidean norm. So, the Hausdorff distance between $M_i$ and $M_t$, denoted by $d(M_i, M_t)$, is defined as:
\( 
d(M_i, M_t) = \max_{p \in M_i} d(p, M_t)
\). 
Here, $d(M_i, M_t)$ is the forward distance and to $d(M_t, M_i)$ is the backward distance.  
So, the Hausdorff Distance,
\( 
d_{HD}(M_i, M_t) = \max \big( d(M_i, M_t), \; d(M_t, M_i) \big).
\)

\textbf{Normal Consistency (NC):} This metric measures the angular alignment between the reconstructed and ground-truth surface normals~\cite{mescheder2019occupancy}. 
\( NC = \frac{1}{N} \sum_{i=1}^{N} \left| n_i^s \cdot n^t_{nearest-neighbor(i)} \right| \), 
where $n_i^s$ represents the normal at the $i$-th sample point on the reconstructed surface and  $n^t_{nearest-neighobr(i)}$ is the normal at its nearest neighbor on the ground-truth surface. %

\textbf{Relative Error (RE):}
We compute the relative error between the area of the reconstructed leaf surface, $\mathcal{A}_R$, and the ground-truth $\mathcal{A}_G$ to assess leaf area accuracy:
\(RE = \frac{|\mathcal{A}_R- \mathcal{A}_G|}{ \mathcal{A}_G} \times 100\).  

\subsection{Leaf Point Cloud Coverage}
\begin{figure}[!htbp]
    \centering
    \includegraphics[width=1.0\linewidth]{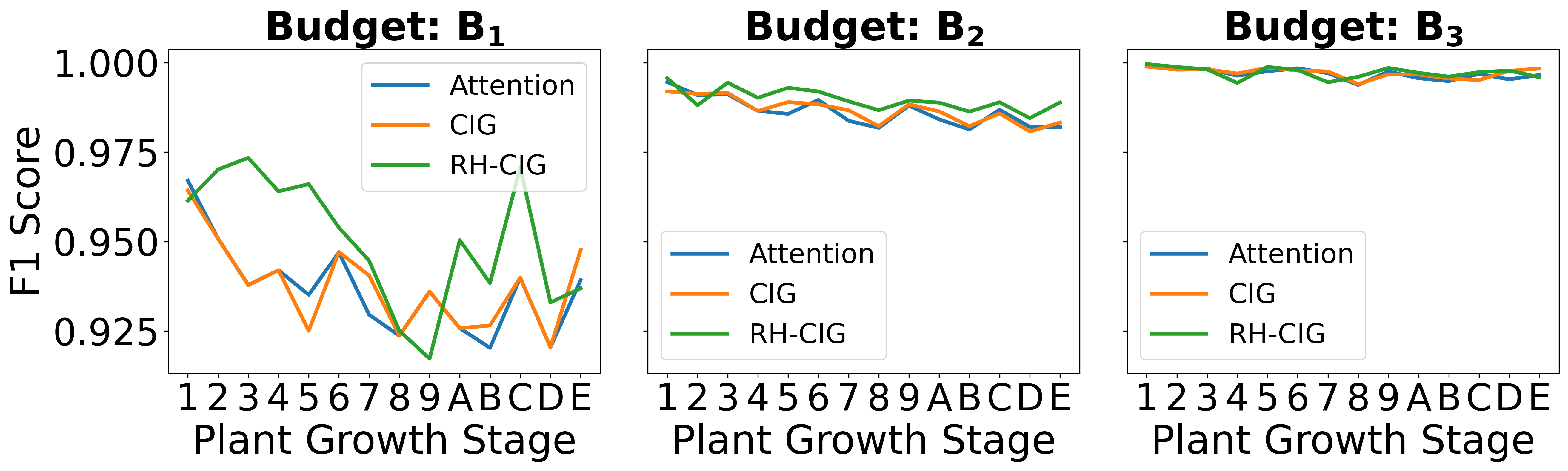}
    \caption{F1-scores averaged over both plants for CIG and RH-CIG planners with respect to Attention Benchmark, at Travel Budgets: $B_1$ (left),  $B_2$ (center), and  $B_3$ (right). 
    }
    \label{fig:f1_scores}
\end{figure}

Fig.~\ref{fig:f1_scores} shows F1-score progression across the 14 growth stages averaged over both plants (A2 and B1) at three budget levels. Table-\ref{tab:f1-score} presents detailed per-plant coverage results.

At budget $B_1$, RH-CIG achieves a mean F1-score of $95.04\%$ compared to $93.67\%$ for Attention-NBV and $93.77\%$ for CIG-NBV. RH-CIG outperforms both baselines on 21 of 28 plants ($75\%$). Across both plants, 20 of 28 plant-growth-stage combinations  produce near-identical F1 scores between Attention-NBV and CIG-NBV, with CIG-NBV recording a higher value in only 5 cases and Attention-NBV in 3.

At earlier growth stages, where most leaves are directly visible, all planners produce comparable coverage. RH-CIG improves reconstruction quality at later growth stages where inter-leaf occlusion increases, with representative improvements at A22 $(98.8\% >95.1\%)$, B13 $(97.2\%>93.5\%)$, B15 $(95.8\%> 92.3\%)$, B1A $(93.2\%>89.6\%)$, and B1C $(96.3\%>93.2\%)$ (Table-\ref{tab:f1-score}). At B19 and B1E, RH-CIG-NBV records the lowest F1-score among the three methods.  

At budget $B_2$, RH-CIG outperforms Attention-NBV and CIG-NBV in 22 and 25 cases, respectively, with all methods exceeding $98.5\%$ (Fig.~\ref{fig:f1_scores}); RH-CIG achieves the highest F1-score ($98.97\%$). 
At budget $B_3$, RH-CIG outperforms the other methods on 17 plants, while Attention-NBV and CIG-NBV achieve higher coverage on 4 plants each, with 3 ties. 
All methods converge to near-complete coverage $(99.7\%)$, indicating budget saturation.
Overall, RH-CIG remains competitive under tighter travel budgets where efficient viewpoint selection is important.

\begin{table*}[htbp]
\centering
\caption{Leaf Surface Reconstruction Performance of CIG and RH-CIG Planning methods with respect to Attention Benchmark  
}
\label{tab:qualitative} 
\renewcommand{\arraystretch}{1.3} 
\begin{tabular}{ l|cc|cc|cc}
\hline
\hline 

& \multicolumn{2}{c|}{\textbf{Hausdorff Distance}  } 

& \multicolumn{2}{c|}{\textbf{RMS (mesh to mesh)}  }
& \multicolumn{2}{c}{\textbf{Relative  Error (area) }} \\
\cline{2-7}

\multicolumn{1}{c|}{\multirow{2}{*}[10pt]{  \textbf{Plant ID }}}
 & CIG(\%)   & RH-CIG(\%)   
  & CIG(\%)   & RH-CIG(\%) 
 & CIG(\%)   & RH-CIG(\%)  \\
\hline 

A2\_20220512\_a (A21)   & -0.88 & -1.80 &   +\underline{0.05} & -0.22 &   -0.08 & -1.12 \\
A2\_20220519\_a (A22)   & -0.11 & +\underline{7.68}   & -0.40 & +\underline{6.23} &   +0.00 & +\underline{2.47} \\
A2\_20220525\_a (A23)   & +1.01 & +\underline{9.87}  & +0.68 & +\underline{7.59} &   +\underline{0.28} & +0.20 \\
A2\_20220531\_a (A24)   & -0.90 & +\underline{8.05}  & +0.34 & +\underline{9.43} &   +0.00 & -0.36 \\
A2\_20220608\_a (A25)   & +0.14 & +\underline{6.51}   & -4.25 & +\underline{1.69} &   -2.19 & -1.14 \\
A2\_20220616\_a (A26)   & +\underline{0.97} & +0.63   & -1.51 & +\underline{1.86} &   -0.41 & +\underline{0.72} \\
A2\_20220620\_a (A27)   & +\underline{2.14} & +1.64   & -0.38 & +\underline{1.23} &   -0.16 & +\underline{0.27} \\
A2\_20220623\_a (A28)   & +1.57 & +\underline{1.98}   & +\underline{0.15} & -4.29 &  +\underline{0.13} & -0.95 \\
A2\_20220627\_a (A29)   & -0.41 & +\underline{0.96}  & +0.83 & +\underline{7.78} &   +0.69 & +\underline{2.87} \\
A2\_20220629\_a (A2A)   & +0.03 & +\underline{1.29}   & -0.73 & +\underline{4.17} &   +0.00 & +\underline{0.92} \\
A2\_20220707\_a (A2B)   & -1.91 & +\underline{1.22} &  -3.18 & +\underline{4.85}  & +0.16 & +\underline{1.12} \\
A2\_20220715\_a (A2C)   & +0.19 & +\underline{2.19} &  +0.02 & +\underline{2.51}   & -0.04 & +\underline{0.08} \\
A2\_20220721\_a (A2D)  & -0.95 & +\underline{4.22} &   -0.93 & +\underline{11.89}   & -0.45 & +\underline{2.03} \\
A2\_20220729\_a (A2E)  & +7.00 & +\underline{9.65} &   +16.89 & +\underline{22.99} &  +2.14 & +\underline{2.90} \\
\hline 
\multicolumn{1}{r|}{\textbf{Mean}}   &  +0.56 & +\underline{3.86} &   +0.54 & +\underline{5.55} &   +0.005 & +\underline{0.71} \\
\hline 
\hline 
B1\_20220513\_a (B11)  & -5.62 & +\underline{1.69} &   -1.70 & +\underline{2.56} &  -1.50 & -1.63 \\
B1\_20220520\_a (B12)  & -0.68 & +\underline{1.36} &   +\underline{2.28} & -0.48 &  +\underline{0.96} & +0.44 \\
B1\_20220526\_a (B13)   & -0.64 & +\underline{8.57} &   -0.21 & +\underline{18.72} &   +0.08 & +\underline{1.76} \\
B1\_20220531\_a (B14)   & +0.51 & +\underline{5.00} &   -0.56 & +\underline{1.09} &   -0.00 & +\underline{1.55} \\
B1\_20220609\_a (B15)   & -0.30 & -1.99   & -0.59 & +\underline{4.23} &  -0.92 & +\underline{3.02} \\
B1\_20220617\_a (B16)   & +0.02 & +\underline{2.73} &   +\underline{0.16} & -1.21  & -0.00 & +\underline{0.06} \\
B1\_20220620\_a (B17)   & +\underline{8.42} & +7.84 &   +\underline{3.46} & -4.48 &   -0.44 & -1.48 \\
B1\_20220623\_a (B18)   & +0.63 & +\underline{0.88} &  +\underline{0.11} & -1.66 &   -0.00 & +\underline{1.34} \\
B1\_20220627\_a (B19)  & -0.27 & -4.07   & -1.30 & -8.10  & +\underline{0.21} & -1.87 \\
B1\_20220630\_a (B1A)  & +0.36 & +\underline{3.95} &   +0.34 & +\underline{3.55} &   +0.07 & +\underline{1.59} \\
B1\_20220707\_a (B1B)   & +1.42 & +\underline{7.57} &  -3.50 & +\underline{9.30} &  -1.19 & -0.14 \\
B1\_20220715\_a (B1C)  & -1.56 & +\underline{2.27} &   -3.34 & +\underline{8.26} &  -0.88 & +\underline{1.12} \\
B1\_20220721\_a (B1D)   & +1.12 & +\underline{4.47} &   +2.05 & +\underline{9.27} &  +0.05 & +\underline{3.70} \\
B1\_20220729\_a (B1E)   & +\underline{9.38} & +1.32 & +\underline{7.50} & -2.68 & -0.05 & +\underline{0.79} \\
\hline 
\multicolumn{1}{r|}{\textbf{Mean}} &  +0.91 & +\underline{2.97}   	& +0.34 &	+\underline{2.74} &   	-0.25 	& +\underline{0.73} \\
\hline
\hline 
\addlinespace
\multicolumn{7}{c}{ \underline{Underline}: denotes best value. Results are shown as $\%$ improvement w.r.t Benchmark (Attention-NBV) \cite{burusa2024attention}. }

\end{tabular}

\end{table*}

\subsection{Leaf Surface Reconstruction Quality}

In Table~\ref{tab:qualitative}, we report per-leaf HD, RMS, and RE averaged across all budget levels ($B_1, B_2, B_3 $)  for both plants. 
The downstream surface reconstruction evaluation reveals differences between planning strategies that are not visible at the F1 coverage level, motivating its use as the primary evaluation criterion for phenotyping applications.

CIG achieves a $0.74\%$ mean HD improvement over Attention-NBV, which validates that geometric awareness improves surface quality. 
RH-CIG amplifies this to a mean improvement of $+3.42\%$ (IQR:$1.27\%-6.77\%$). Improvements are most significant at mid-to-late growth stages ($+9.86\%$). 
For early-stage plants, RH-CIG shows comparable or slightly worse HD, suggesting limited benefit for minimal occlusion cases. 
For NC, all methods achieve similar values, ranging from 0.45 to 0.62. RH-CIG shows marginal improvements on 11 of 28 plants, indicating that surface normal accuracy is predominantly constrained by point density and that our CIG formulation penalizes clustered points.  
RH-CIG achieves lower RMS than Attention-NBV on 21 of 28 plants ($75\%$) at mid-to-late growth stages.

\begin{figure}[!htbp]
    \centering
    \includegraphics[width=1.0\columnwidth]{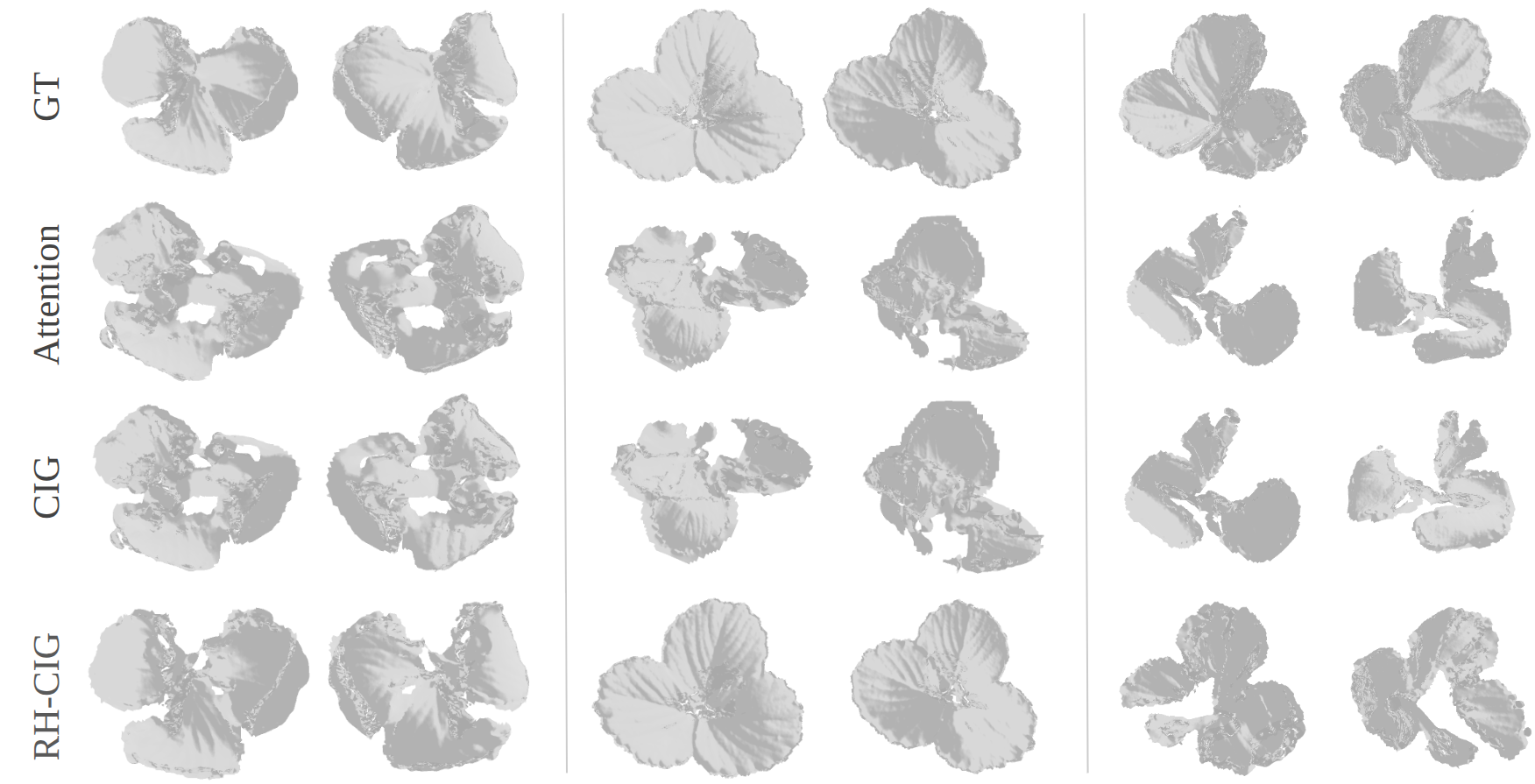}
    \caption{Ground truth vs reconstructed leaf surfaces after view planning.  
    }
    \label{fig:qual_surface}
\end{figure}

Moreover, upon further inspection of B11, the largest single contributor to Attention-NBV's advantage across all surface metrics for plant B1, we observed an anomalous reconstruction case with the CIG-NBV method.
One leaf instance contained only a single reconstructed point, making it impossible for Poisson surface reconstruction to produce a geometrically valid mesh for that leaf. 
Consequently, the downstream surface evaluation metrics for this stage reflect the reconstruction failure, with differences of $-5.62\%$ HD, $-1.70\%$ RMS, and $-1.50\%$ RE. In contrast, the F1 score shows only a minor difference ($95.47\%$  vs $94.92\%$), indicating limited sensitivity to this failure. 
Excluding this growth stage, CIG-NBV consistently reduces HD, further supporting the conclusion that CIG has a genuine downstream effect on leaf surface geometry recovery that Attention-NBV does not achieve. 
Across all budget levels, the methods execute 12-18 views within travel budget constraints and achieve more than $ 95\%$ budget utilization, confirming that performance differences stem from viewpoint geometry selection rather than differing view counts or incomplete budget use. 
Fig.~\ref{fig:qual_surface} presents representative results for trifoliate strawberry leaves from LAST-STRAW using CIG and RH-CIG compared with the Attention-NBV benchmark.
The ground-truth leaves show smooth surfaces and complete coverage. 
The CIG and Attention view-planning methods produce fragmented reconstructions with significant gaps or missing leaflets. 
Our RH-CIG method produces the most complete reconstructions, maintaining edge features while achieving smooth surface continuity. 
Visual inspection reveals the benefit of our CIG function and receding-horizon NBV planning compared to the benchmark~\cite{burusa2024attention}.

\section{Conclusion and Future Work}
In this paper, we develop a receding-horizon next-best-view planning method for autonomous leaf surface reconstruction and present simulation results using a real strawberry dataset. The evaluation shows that our novel centroid-based information gain function can capture the spatial geometry of leaves better than a coverage-based utility function. We use Poisson-based surface reconstruction for downstream reconstruction. Future extensions of this work involve the design of lighter reconstruction methods similar to \cite{zermas2017estimating} for real-time deployment \cite{ahmed2026review}. To better handle failure cases such as B19, future work will investigate incorporating an RL-based adaptive strategy \cite{ruckin2022adaptive} within our RH-CIG NBV planner framework.
In addition, we plan to validate our proposed approach through real-world UAV field experiments to evaluate its performance.





\balance
\bibliographystyle{IEEEbib}
\bibliography{iros_ref.bib}

@article{ahmed2026review,
  title={Review and Evaluation of Point-Cloud based Leaf Surface Reconstruction Methods for Agricultural Applications},
  author={Ahmed, Arif and Maini, Parikshit},
  journal={arXiv preprint arXiv:2604.03328},
  year={2026}
}

@inproceedings{aspert2002mesh,
  title={Mesh: Measuring errors between surfaces using the hausdorff distance},
  author={Aspert, Nicolas and Santa-Cruz, Diego and Ebrahimi, Touradj},
  booktitle={Proceedings. IEEE international conference on multimedia and expo},
  volume={1},
  pages={705--708},
  year={2002},
  organization={IEEE}
}

@inproceedings{bengio2009curriculum,
  title={Curriculum learning},
  author={Bengio, Yoshua and Louradour, J{\'e}r{\^o}me and Collobert, Ronan and Weston, Jason},
  booktitle={Proceedings of the 26th annual international conference on machine learning},
  pages={41--48},
  year={2009}
}

@inproceedings{bircher2016receding,
  title={Receding horizon" next-best-view" planner for 3d exploration},
  author={Bircher, Andreas and Kamel, Mina and Alexis, Kostas and Oleynikova, Helen and Siegwart, Roland},
  booktitle={2016 IEEE international conference on robotics and automation (ICRA)},
  pages={1462--1468},
  year={2016},
  organization={IEEE}
}

@article{bircher2018receding,
  title={Receding horizon path planning for 3D exploration and surface inspection},
  author={Bircher, Andreas and Kamel, Mina and Alexis, Kostas and Oleynikova, Helen and Siegwart, Roland},
  journal={Autonomous Robots},
  volume={42},
  number={2},
  pages={291--306},
  year={2018},
  publisher={Springer}
}

@article{burusa2024attention,
  title={Attention-driven next-best-view planning for efficient reconstruction of plants and targeted plant parts},
  author={Burusa, Akshay K and van Henten, Eldert J and Kootstra, Gert},
  journal={Biosystems Engineering},
  volume={246},
  pages={248--262},
  year={2024},
  publisher={Elsevier}
}

@inproceedings{burusa2024gradient,
  title={Gradient-based local next-best-view planning for improved perception of targeted plant nodes},
  author={Burusa, Akshay K and van Henten, Eldert J and Kootstra, Gert},
  booktitle={2024 IEEE International Conference on Robotics and Automation (ICRA)},
  pages={15854--15860},
  year={2024},
  organization={IEEE}
}

@inproceedings{dhami2023pred,
  title={Pred-nbv: Prediction-guided next-best-view planning for 3d object reconstruction},
  author={Dhami, Harnaik and Sharma, Vishnu D and Tokekar, Pratap},
  booktitle={2023 IEEE/RSJ International Conference on Intelligent Robots and Systems (IROS)},
  pages={7149--7154},
  year={2023},
  organization={IEEE}
}

@inproceedings{dhami2024map,
  title={Map-nbv: Multi-agent prediction-guided next-best-view planning for active 3d object reconstruction},
  author={Dhami, Harnaik and Sharma, Vishnu Dutt and Tokekar, Pratap},
  booktitle={2024 IEEE/RSJ International Conference on Intelligent Robots and Systems (IROS)},
  pages={5724--5731},
  year={2024},
  organization={IEEE}
}

@inproceedings{isaacjose2025iros,
  title={{GO-VMP}: {G}lobal Optimization for View Motion Planning in Fruit Mapping},
  author={Isaac Jose, Allen and Pan, Sicong and Zaenker, Tobias and Menon, Rohit and Houben, Sebastian and Bennewitz, Maren},
  booktitle={IEEE/RSJ International Conference on Intelligent Robots and Systems (IROS)},
  year={2025}
}

@article{jia2025pb,
  title={PB-NBV: Efficient Projection-Based Next-Best-View Planning Framework for Reconstruction of Unknown Objects},
  author={Jia, Zhizhou and Li, Yuetao and Hao, Qun and Zhang, Shaohui},
  journal={IEEE Robotics and Automation Letters},
  year={2025},
  publisher={IEEE}
}

@incollection{katz2007direct,
  title={Direct visibility of point sets},
  author={Katz, Sagi and Tal, Ayellet and Basri, Ronen},
  booktitle={ACM SIGGRAPH 2007 papers},
  pages={24--es},
  year={2007}
}

@inproceedings{kazhdan2006poisson,
  title={Poisson surface reconstruction},
  author={Kazhdan, Michael and Bolitho, Matthew and Hoppe, Hugues},
  booktitle={Proceedings of the fourth Eurographics symposium on Geometry processing},
  volume={7},
  number={4},
  year={2006}
}

@article{knapitsch2017tanks,
  title={Tanks and temples: Benchmarking large-scale scene reconstruction},
  author={Knapitsch, Arno and Park, Jaesik and Zhou, Qian-Yi and Koltun, Vladlen},
  journal={ACM Transactions on Graphics (ToG)},
  volume={36},
  number={4},
  pages={1--13},
  year={2017},
  publisher={ACM New York, NY, USA}
}

@inproceedings{kuffner2000rrt,
  title={RRT-connect: An efficient approach to single-query path planning},
  author={Kuffner, James J and LaValle, Steven M},
  booktitle={Proceedings 2000 ICRA. Millennium conference. IEEE international conference on robotics and automation. Symposia proceedings (Cat. No. 00CH37065)},
  volume={2},
  pages={995--1001},
  year={2000},
  organization={IEEE}
}

@article{lin2021uav,
  title={UAV based estimation of forest leaf area index (LAI) through oblique photogrammetry},
  author={Lin, Lingchen and Yu, Kunyong and Yao, Xiong and Deng, Yangbo and Hao, Zhenbang and Chen, Yan and Wu, Nankun and Liu, Jian},
  journal={remote sensing},
  volume={13},
  number={4},
  pages={803},
  year={2021},
  publisher={MDPI}
}

@article{liu2023fast,
  title={Fast reconstruction method of three-dimension model based on dual RGB-D cameras for peanut plant},
  author={Liu, Yadong and Yuan, Hongbo and Zhao, Xin and Fan, Caihu and Cheng, Man},
  journal={Plant Methods},
  volume={19},
  number={1},
  pages={17},
  year={2023},
  publisher={Springer}
}

@inproceedings{menon2023nbv,
  title={NBV-SC: Next best view planning based on shape completion for fruit mapping and reconstruction},
  author={Menon, Rohit and Zaenker, Tobias and Dengler, Nils and Bennewitz, Maren},
  booktitle={2023 IEEE/RSJ international conference on intelligent robots and systems (IROS)},
  pages={4197--4203},
  year={2023},
  organization={IEEE}
}

@inproceedings{mescheder2019occupancy,
  title={Occupancy networks: Learning 3d reconstruction in function space},
  author={Mescheder, Lars and Oechsle, Michael and Niemeyer, Michael and Nowozin, Sebastian and Geiger, Andreas},
  booktitle={Proceedings of the IEEE/CVF conference on computer vision and pattern recognition},
  pages={4460--4470},
  year={2019}
}

@book{rogers1998hausdorff,
  title={Hausdorff measures},
  author={Rogers, Claude Ambrose},
  year={1998},
  publisher={Cambridge University Press}
}

@inproceedings{roy2017active,
  title={Active view planning for counting apples in orchards},
  author={Roy, Pravakar and Isler, Volkan},
  booktitle={2017 IEEE/RSJ International Conference on Intelligent Robots and Systems (IROS)},
  pages={6027--6032},
  year={2017},
  organization={IEEE}
}

@inproceedings{ruckin2022adaptive,
  title={Adaptive informative path planning using deep reinforcement learning for uav-based active sensing},
  author={R{\"u}ckin, Julius and Jin, Liren and Popovi{\'c}, Marija},
  booktitle={2022 International Conference on Robotics and Automation (ICRA)},
  pages={4473--4479},
  year={2022},
  organization={IEEE}
  }

@inproceedings{yu2021pointr,
  title={Pointr: Diverse point cloud completion with geometry-aware transformers},
  author={Yu, Xumin and Rao, Yongming and Wang, Ziyi and Liu, Zuyan and Lu, Jiwen and Zhou, Jie},
  booktitle={Proceedings of the IEEE/CVF international conference on computer vision},
  pages={12498--12507},
  year={2021}
}

@inproceedings{li2025semp,
  title={SemP-NBV: Semantic-Aware Predictive Next-Best-View for Autonomous Plant 3D Reconstruction},
  author={Li, Xingjian and He, Weilong and Park, Jeremy and Reberg-Horton, Chris and Mirsky, Steven and Lobaton, Edgar and Xiang, Lirong},
  booktitle={2025 IEEE/RSJ International Conference on Intelligent Robots and Systems (IROS)},
  pages={1210--1216},
  year={2025},
  organization={IEEE}
}

@article{tunca2024accurate,
  title={Accurate leaf area index estimation in sorghum using high-resolution UAV data and machine learning models},
  author={Tunca, Emre and K{\"o}ksal, Ey{\"u}p Selim and {\"O}zt{\"u}rk, Elif and Akay, Hasan and Taner, Sakine {\c{C}}etin},
  journal={Physics and Chemistry of the Earth, Parts A/B/C},
  volume={133},
  pages={103537},
  year={2024},
  publisher={Elsevier}
}

@misc{uol2024laststraw,
   title = {{Lincoln's Annotated Spatio-Temporal Strawberry Dataset (LAST-Straw)}}, 
   author = {James, Katherine Margaret Frances and Heiwolt, Karoline and Sargent, Daniel James and Cielniak, Grzegorz},
   year = {2024},
   eprint = {2403.00566},
   archivePrefix = {arXiv},
   primaryClass = {cs.CV}
}

@article{wei2024fast,
  title={Fast Multi-View 3D reconstruction of seedlings based on automatic viewpoint planning},
  author={Wei, Kaihua and Liu, Shuang and Chen, Qingguang and Huang, Shentao and Zhong, Mingwei and Zhang, Jingcheng and Sun, Hongwei and Wu, Kaihua and Fan, Shanhui and Ye, Ziran and others},
  journal={Computers and Electronics in Agriculture},
  volume={218},
  pages={108708},
  year={2024},
  publisher={Elsevier}
}

@article{wu2022miniaturized,
  title={A miniaturized phenotyping platform for individual plants using multi-view stereo 3D reconstruction},
  author={Wu, Sheng and Wen, Weiliang and Gou, Wenbo and Lu, Xianju and Zhang, Wenqi and Zheng, Chenxi and Xiang, Zhiwei and Chen, Liping and Guo, Xinyu},
  journal={Frontiers in plant science},
  volume={13},
  pages={897746},
  year={2022},
  publisher={Frontiers Media SA}
}

@article{yuan2023field,
  title={Field phenotyping monitoring systems for high-throughput: A survey of enabling technologies, equipment, and research challenges},
  author={Yuan, Huali and Song, Minghan and Liu, Yiming and Xie, Qi and Cao, Weixing and Zhu, Yan and Ni, Jun},
  journal={Agronomy},
  volume={13},
  number={11},
  pages={2832},
  year={2023},
  publisher={MDPI}
}

@inproceedings{zeng2020pc,
  title={Pc-nbv: A point cloud based deep network for efficient next best view planning},
  author={Zeng, Rui and Zhao, Wang and Liu, Yong-Jin},
  booktitle={2020 IEEE/RSJ International Conference on Intelligent Robots and Systems (IROS)},
  pages={7050--7057},
  year={2020},
  organization={IEEE}
}

@inproceedings{zermas2017estimating,
  title={Estimating the leaf area index of crops through the evaluation of 3D models},
  author={Zermas, Dimitris and Morellas, Vassilios and Mulla, David and Papanikolopoulos, Nikolaos},
  booktitle={2017 IEEE/RSJ International Conference on Intelligent Robots and Systems (IROS)},
  pages={6155--6162},
  year={2017},
  organization={IEEE}
}

@article{zhang2012application,
  title={The application of small unmanned aerial systems for precision agriculture: a review},
  author={Zhang, Chunhua and Kovacs, John M},
  journal={Precision agriculture},
  volume={13},
  number={6},
  pages={693--712},
  year={2012},
  publisher={Springer}
}

@article{zhang2024improved,
  title={Improved estimation of cotton (Gossypium hirsutum L.) LAI from multispectral data using UAV point cloud data},
  author={Zhang, Lechun and Sun, Binshu and Zhao, Denan and Shan, Changfeng and Wang, Baoju and Wang, Guobin and Song, Cancan and Chen, Pengchao and Lan, Yubin},
  journal={Industrial Crops and Products},
  volume={217},
  pages={118851},
  year={2024},
  publisher={Elsevier}
}

\end{document}